\documentclass[conference]{IEEEtran}
\IEEEoverridecommandlockouts
\usepackage{cite}
\usepackage{amsmath,amssymb,amsfonts}
\usepackage{algorithmic}
\usepackage{graphicx}
\usepackage{glossaries}
\usepackage{multirow}
\usepackage{subcaption}
\usepackage{textcomp}
\usepackage{xcolor}
\usepackage{makecell}
\def\BibTeX{{\rm B\kern-.05em{\sc i\kern-.025em b}\kern-.08em
    T\kern-.1667em\lower.7ex\hbox{E}\kern-.125emX}}

\newacronym{nlp}{NLP}{Natural Language Processing}
\newacronym{ec}{EC}{Evolutionary Computation}
\newacronym{ea}{EA}{Evolutionary Algorithm}
\newacronym{es}{ES}{Evolutionary Strategies}
\newacronym{dl}{DL}{Deep Learning}
\newacronym{cnn}{CNN}{Convolutional Neural Network}
\newacronym{dnn}{DNN}{Deep Neural Network}
\newacronym{ssl}{SSL}{Self-Supervised learning}
\newacronym{ne}{NE}{Neuroevolution}
\newacronym{ae}{AE}{Auto-Encoder}
\newacronym{vae}{VAE}{Variational Auto Encoder}
\newacronym{gan}{GAN}{Generative Adversarial Network}
\newacronym{dsge}{DSGE}{Dynamic Structured Grammatical Evolution}

\begin{document}

\title{Towards evolution of Deep Neural Networks through contrastive Self-Supervised learning}


\author{\IEEEauthorblockN{Adriano Vinhas}
\IEEEauthorblockA{\textit{University of Coimbra} \\
\textit{CISUC/LASI, DEI}\\
Coimbra, Portugal \\
avinhas@dei.uc.pt}
\and
\IEEEauthorblockN{João Correia}
\IEEEauthorblockA{\textit{University of Coimbra} \\
\textit{CISUC/LASI, DEI}\\
Coimbra, Portugal \\
jncor@dei.uc.pt}
\and
\IEEEauthorblockN{Penousal Machado}
\IEEEauthorblockA{\textit{University of Coimbra} \\
\textit{CISUC/LASI, DEI}\\
Coimbra, Portugal \\
machado@dei.uc.pt}
}
\maketitle

\begin{abstract}
~\glspl{dnn} have been successfully applied to a wide range of problems. However, two main limitations are commonly pointed out. The first one is that they require long time to design. The other is that they heavily rely on labelled data, which can sometimes be costly and hard to obtain. In order to address the first problem, neuroevolution has been proved to be a plausible option to automate the design of~\glspl{dnn}. As for the second problem, self-supervised learning has been used to leverage unlabelled data to learn representations. Our goal is to study how neuroevolution can help self-supervised learning to bridge the gap to supervised learning in terms of performance. In this work, we propose a framework that is able to evolve deep neural networks using self-supervised learning. Our results on the CIFAR-10 dataset show that it is possible to evolve adequate neural networks while reducing the reliance on labelled data. Moreover, an analysis to the structure of the evolved networks suggests that the amount of labelled data fed to them has less effect on the structure of networks that learned via self-supervised learning, when compared to individuals that relied on supervised learning.
\end{abstract}


\section{Introduction}
\label{sec:intro}
~\gls{dl} has demonstrated its power by achieving new state of the art results and surpassing previous ones based on more traditional methods. Part of this success is due to the availability of large-scale labelled datasets. However, labels are often time-consuming and costly to obtain. Additionally, one could argue that label-based learning can be limiting, as the produced labels are not inherent in the data and, therefore, do not accurately relate to how animals, including humans, tend to learn. For instance, during early stages of life, young children can learn to discriminate between common categories (e.g. birds and dogs~\cite{babies_birds_vs_dogs}) without being explicitly told that they have different labels. This encourages the idea that, at initial stages, learning is unsupervised and uses perceptual abilities to build representations of those categories~\cite{ssl_child}.

Inspired on this rationale, \gls{ssl} has been growing on popularity, due to its ability to leverage vast amounts of unlabelled data available. Its goal is to learn helpful representations without requiring extrinsic labels. It was found that \glspl{dnn} trained using \gls{ssl} obtained breakthrough results in fields like \gls{nlp}, overshadowing the ones obtained with supervised learning. However, the same progress is yet to be achieved in computer vision. We believe that there are two main reasons for this. The first one is that most of the works that apply \gls{ssl} on computer vision have used a limited range of network topologies. This is because the design of a \gls{dnn} is a time-consuming task, as its hyperparameters are optimised following a trial-and-error process. The second reason is because \gls{ssl} algorithms are manually designed too, which hampers the discovery of new approaches that might be more suited to the image domain.

\gls{ec} is another biologically inspired field which borrows ideas from evolution theory. It has been used to automate the search for the best topology or other hyperparameters of a given \gls{dnn}. The intersection of these two fields originated a subfield known as \gls{ne}. Not only \gls{ne} helps designing the network that better suits a task automatically, it also promotes the emergence of increasingly better solutions by following a bio-inspired approach.

In this paper, we hypothesise that combining \gls{ssl} algorithms with \gls{ne} promotes the emergence of \glspl{dnn} that are useful for specific tasks (in our case, image classification). By merging these two areas, we believe that we can take the best of \gls{ssl} and \gls{ne}: reduce the need for labelled data and, at the same time, automate design aspects that impact the final \glspl{dnn}. In order to achieve this, we propose a framework that performs \textbf{Evo}lution of \textbf{De}ep \textbf{N}etworks through \textbf{S}elf \textbf{S}upervision (EvoDeNSS). The code is publicly available on GitHub\footnote{https://github.com/adrianovinhas/evodenss}.

The remainder of the document is structured as follows. In Section~\ref{sec:related_work} we survey related works. Next, in Section~\ref{sec:proposed_approach}
we detail the developed approach, which is then followed by the undertaken
experiments and respective results (Section~\ref{sec:experimentation}). To end, in Section~\ref{sec:conclusions}, conclusions
are drawn and future work is addressed.

\section{Related work}
\label{sec:related_work}

Our work intersects two main topics - 1)~\gls{ssl} and 2)~\gls{ne}. Therefore, we will start by outlining the dynamics of this learning paradigm, followed by reviewing~\gls{ne} works which attempt to learn representations without the use of extrinsic labels.

\glsreset{ssl}
\subsection{\gls{ssl}}
\label{subsec:ssl}

~\gls{ssl}~\cite{de1994learning, doersch2015unsupervised} is a learning paradigm that uses the input data itself to create pseudo-labels that will guide the learning process. Pseudo-labels are intrinsically generated from the relation between the original input and a modified version of it. This modification can come in the form of parts of the original input, a corrupted version of the input or even different modalities of data. The goal of \gls{ssl} is to create feature extractors by predicting the original inputs based on their modified counterparts.

\begin{figure}[!ht]
    \centering
    \includegraphics[width=0.95\linewidth]{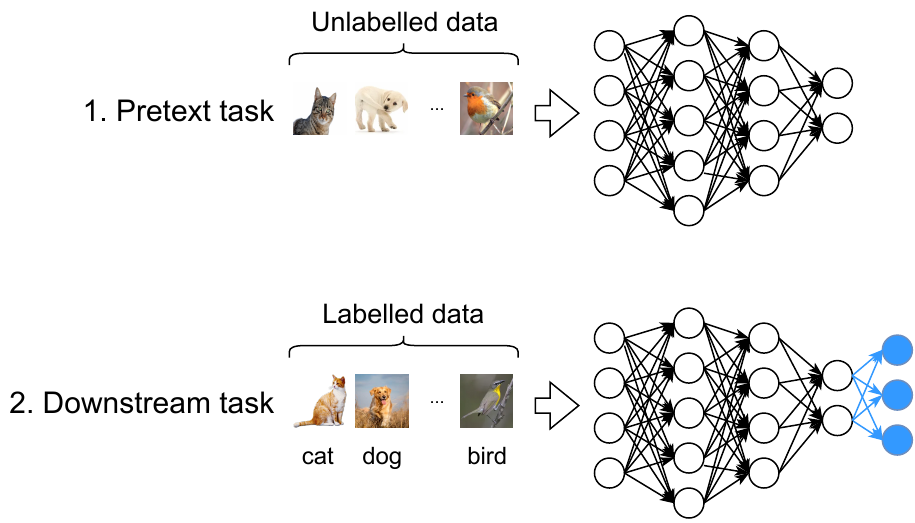}
    \caption{Overview of the \gls{ssl} process for an image classification problem.}
    \label{fig:ssl_process}
\end{figure}

\gls{ssl} methods are composed of two stages. In the first one, we train a model without relying on any labels and without explicitly taking into account our final goal. Instead, the model is trained to solve a task related to the one we are interested in, known as pretext task, that will learn how to extract features based on pseudo-labels. The second stage, known as the downstream task, transfers the pretrained model by reusing the learned features in order to train another model that solves the actual task. This step follows a supervised learning setting, as the features extracted from labelled inputs are used to train the downstream task. An overview of the 2-step process in \gls{ssl} is depicted in Figure~\ref{fig:ssl_process}.


\glsreset{ne}
\subsection{\gls{ne}}
\label{subsec:ne}

\gls{ne}~\cite{floreano2008neuroevolution, montana1989training, harp1991genetic} is a field that encompasses the application of \glspl{ea} to automate design aspects of \glspl{dnn}. Within the specific context of \gls{ssl}, \gls{ne} literature can be divided into the evolution of generative and contrastive methods. 

Regarding generative methods, David and Greental evolved the weights of an \gls{ae}~\cite{david2014genetic}. Their architecture is fixed, and the weights of the encoder are tied to the decoder, allowing to substantially reduce the search space. Lander and Shang~\cite{lander2015evoae} proposed a framework which allows more freedom to the evolution process by targeting the weight and structural space. Assunção et al. evolved \glspl{ae} that are not constrained to have the same structure nor weights on the encoder and decoder~\cite{assuncao2018automatic}, aiming to compress the data without compromising classification performance. Their fitness function is the aggregation of several objectives into a single one via linear combination: 1) maximise the accuracy of the learned representations in an image classification task, 2) minimise the size of the chokepoint, 3) minimise the number of layers in the decoder. Other works employed \gls{ne} to the evolution of \glspl{vae}~\cite{DBLP:journals/corr/KingmaW13}. For instance, Hajewski and Oliveira evolve \glspl{vae} using only fully connected layers~\cite{hajewski2020evolutionary}. The architecture of their evolved \glspl{dnn} was flexible enough to vary the number of layers and the number of neurons within a layer. Chen et al. employed a block-based search by allowing any arrangement of layers within each block (without skip connections)~\cite{chen2020evolving}, but the way blocks interact which other is predefined. Four blocks are defined in total. Fitness is calculated based on the training loss, which may cause generalisation issues due to overfitting.

\glspl{gan}~\cite{NIPS2014_5ca3e9b1} can be considered as another form of \gls{ssl}, given that a generator $G$ learns representations from a pretext task that is taught to distinguish images that come from the training data or generated from the latent space. E-GAN~\cite{wang2019evolutionary} targeted generators in their evolutionary process and assume that the discriminator will always achieve its optimal performance. Generators are evolved only through mutations and separately trained using three predefined objective functions, generating three new \glspl{gan}. Gonzalez and Miikkulainen evolved loss functions adapted to \glspl{gan}~\cite{gonzalez2021evolving}. Individuals represent coefficients of Taylor expansions and are evolved using an \gls{es} algorithm. Costa et al. approached the evolution of \glspl{gan} as a co-evolution problem~\cite{costa2019coevolution}, whereby generators and discriminators follow a similar representation to DeepNEAT~\cite{miikkulainen:chapter18} and evolved individuals using a mutation operator only. Each \gls{gan} is the result of pairing each discriminator with each generator to calculate the fitness for each individual. This work was later extended to use different fitness functions. One of them is a multiobjective function that takes into account both FID and a novelty search score that measures diversity~\cite{costa2020exploring}. The other used a skill rating score~\cite{costa2020using} based on the Glicko-2 rating system~\cite{glickman2012example}.

Most modern \gls{ssl} methods aim to learn representations by contrastive learning. The idea is to feed a model with multiple views of the same input and encourage it to represent contrasting versions of an input similarly, thereby promoting multi-view invariance. Even though contrastive methods obtain state of the art results, \gls{ne} literature is very scarce in automating design aspects learned with these methods.

In order to automate the design of the data augmentation component, Barrett et al. aimed to evolve the augmentation policy parameters to find the one that produces the best representations~\cite{barrett2023evolutionary}.

Outside the image domain, ELo~\cite{piergiovanni2020evolving} is a network that is trained to minimise a joint loss composed of losses on $M$ modalities, trained on $T$ self-supervised tasks. The final loss is divided into two components. One is the linear combination of all losses $L_{m,t}, m \in M, t \in T$. The other is distillation loss $L_{d}$, which is the result of transferring knowledge from one modality to the main network. Coefficients associated to each one of these losses are evolved through an \gls{es} algorithm. Finally, Wei et al. use \gls{ne} to evolve \gls{dnn} architectures that were trained using a \gls{ssl} algorithm~\cite{wei2021self}. As a fitness metric, the authors rely on a surrogate model that predicts the performance of the individuals.

\section{Proposed approach}
\label{sec:proposed_approach}

In this section we detail the most relevant parts of our proposed neuroevolutionary framework. We start by describing the evolutionary engine, focusing on how the evaluation of individuals was adapted to the \gls{ssl} case. Finally, the dataset partitioning process is specified with regards to both \gls{ssl} and supervised learning.

\subsection{Evolutionary Engine}
\label{subsec:evolutionary_engine}

EvoDeNSS is a neuroevolutionary framework that can be considered as an extension of Fast-DENSER~\cite{10.1007/978-3-030-16670-0_13}. Therefore, we will describe how Fast-DENSER works and what relevant extensions were performed to enable individuals to be evaluated through a \gls{ssl} algorithm.

Fast-DENSER is a \gls{ne} framework that allows the emergence of \glspl{cnn} using a $(1+\lambda)$-\acrshort{es} algorithm. The rationale behind this decision is due to an effort to reduce the number of evaluations, when compared to its predecessor~\cite{assunccao2019denser}. Fast-DENSER aims to run with a more restricted population size, hence limiting the number of evaluations needed to execute the \gls{ne} algorithm.

\begin{figure}[ht]
  \centering
  \includegraphics[width=1.\linewidth]{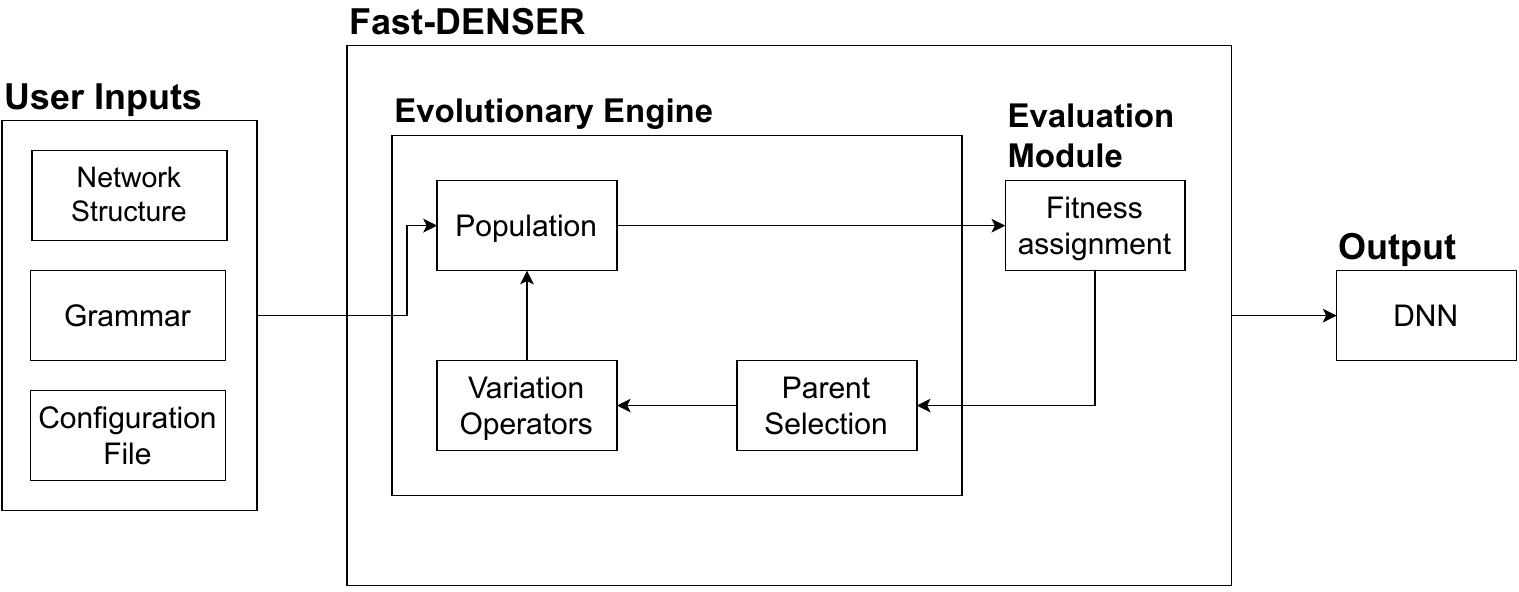}
  \caption{Overview of Fast-DENSER.}
  \label{fig:fast_denser}
\end{figure}

One of the concerns raised by the authors was the impact of this decision in the diversity of the evolved solutions. Diversity is an important factor in \acrshortpl{ea}, as it impacts the ability of converging towards optimal \glspl{dnn}. To compensate this possible diversity loss, weight-sharing mechanisms were not used. Instead, at each generation one parent is selected to create its descendants, but its weights are discarded. In the next generation, the parent is forced to be trained from scratch, using another set of initial weights. An overview of Fast-DENSER is depicted in Figure~\ref{fig:fast_denser}.

EvoDeNSS uses the same representation as Fast-DENSER. It is a two-level representation which can be seen as an array of modules at the macro level (modules can represent a layer or learning aspects), whereas at the microscopic level, each module uses a \gls{dsge} representation~\cite{lourencco2018structured}, which allows the structure of each module to be constrained by context-free grammars. The flexibility of this approach biases the search space towards more favourable solutions by injecting a-priori knowledge. In EvoDeNSS, one does not use fully-connected layers for representation learning, meaning that we only consider layers within the context of learning features. The details about the grammars that were used in EvoDenSS can be checked in the public repository mentioned in Section~\ref{sec:intro}. Moreover, for the learning block we included the possibility of training \glspl{dnn} using LARS optimiser~\cite{you2017large}. This decision was made as this optimiser was chosen by the authors of the \gls{ssl} algorithm we use for evaluation purposes. More details about this algorithm can be found on Section~\ref{subsec:evaluation_module}.

\subsection{Evaluation Module}
\label{subsec:evaluation_module}

In order to evaluate each individual, the genotype needs to be converted into a phenotype, which in this case is a \gls{dnn} in a format that is trainable with a Deep Learning framework. Given that we are focusing on an image classification problem, we define fitness as the ability of an evolved network to correctly classify the images it is fed with. For this effect, we compute fitness using the accuracy on a dedicated subset, created only for this purpose.

\begin{figure}[ht]
    \centering
    \begin{subfigure}[ht]{0.45\linewidth}
        \centering
        \includegraphics[width=1.0\linewidth]{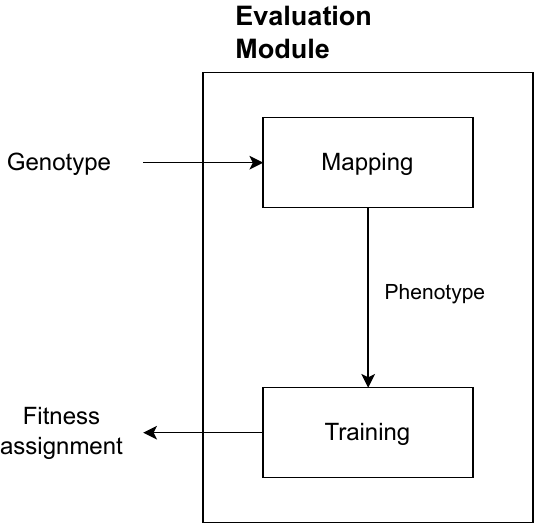}
    \caption{Fitness assignment under supervised learning}
    \label{fig:sl_evaluation}
    \end{subfigure}
    \hspace{0.2cm}
    \begin{subfigure}[ht]{0.45\linewidth}
        \centering
        \includegraphics[width=1.0\linewidth]{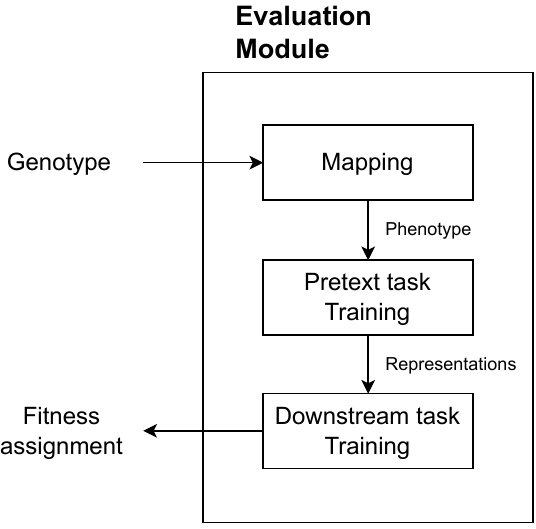}
        \caption{Fitness assignment under \gls{ssl}}
    \label{fig:ssl_evaluation}
    \end{subfigure}
    
    \caption{Overview of individual evaluation under different learning paradigms.}
    \label{fig:evaluation}
\end{figure}

The process to compute accuracy differs slightly depending on the chosen learning paradigm. In supervised learning the model needs to be trained using labelled data (subjected to a maximum training time) and then the accuracy is measured on the dedicated test set. As for \gls{ssl}, the evolved network is trained for the pretext task with unlabelled data to learn representations, bounded by a maximum training time too. Then, the train for the downstream task happens, by applying a linear layer on top of the representations and feeding the network with labelled data. This linear layer is trained with fixed parameters for a predefined number of epochs, meaning that no aspects of the downstream task are incorporated into evolution. In this scenario, the fitness metric is the quality of the learned representations, which could be translated into finding out if the evolved representations are good predictors. Therefore, we measure the accuracy on the downstream task by feeding the dedicated test set data. The differences in calculating fitness for each learning paradigm are depicted in Figure~\ref{fig:evaluation}.

\begin{figure}[ht]
  \centering
  \includegraphics[width=.95\linewidth]{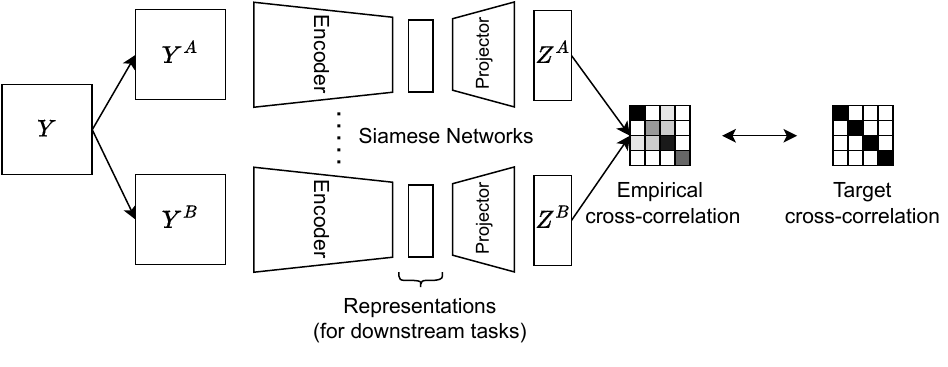}
  \caption{Barlow Twins algorithm}
  \label{fig:bt}
\end{figure}

In order to train representations without using labels, we incorporated Barlow Twins algorithm~\cite{zbontar2021barlow} within the proposed framework. The rationale of this method is to build representations from pairs of image views. Learning occurs by employing a loss function that maximises invariance and minimises the redundancy between these representation pairs. An overview of Barlow Twins method is depicted in Figure~\ref{fig:bt}. 

Each batch of original images from the dataset is augmented, generating two different batches, named $Y^{A}$ and $Y^{B}$. Both $Y^{A}$ and $Y^{B}$ contain unique image views that are created from the same set of original images. Image views are fed to an encoder network, which is in turn attached to a projector network. A projector network is commonly a multiple layer neural network composed by dense and/or batch normalisation layers only. The outputs of this process are two matrices that represent two batches of output vectors $Z^{A}$ and $Z^{B}$. The cross-correlation matrix between $Z^{A}$ and $Z^{B}$ is calculated, and is then used to calculate the loss. The rationale is that we want the same dimensions to be as correlated as possible, making the representations to be invariant to transformations. At the same time different dimensions should be untangled from each other. If they are not, two dimensions can be considered redundant. Therefore, the goal of the optimisation process is to approximate the cross-correlation matrix as much as possible to the identity matrix, via Equation~\ref{eq:bt}.

\begin{equation}
    \label{eq:bt}
    L_{BT} = \underbrace{\sum_{i}(1 - C_{i,i})^2}_\text{invariance term} + \lambda \cdot \underbrace{\sum_{i}\sum_{j} C_{i,j}^{~~2}}_\text{redundancy reduction term},
\end{equation}

In this loss function, $i$ and $j$ identify dimensions of the representation vector, $C$ represents the cross-correlation matrix and $\lambda$ is a parameter that sets the importance of the redundancy reduction term. It should be noted that the representations to be used for downstream tasks are the ones produced by the encoder, hence the projector network is only needed during the pretext task. 

Barlow Twins was chosen for these experiments due to the simplicity of its design when compared to other algorithms. Besides, this algorithm is reported to work well with lower batch sizes when compared to some of its peers, making it a suitable candidate in scenarios that do not require extensive GPU memory resources.

\subsection{Dataset Partitioning}
\label{subsec:dataset_partitioning}

\begin{figure}[ht]
  \centering
  \includegraphics[width=1.\linewidth]{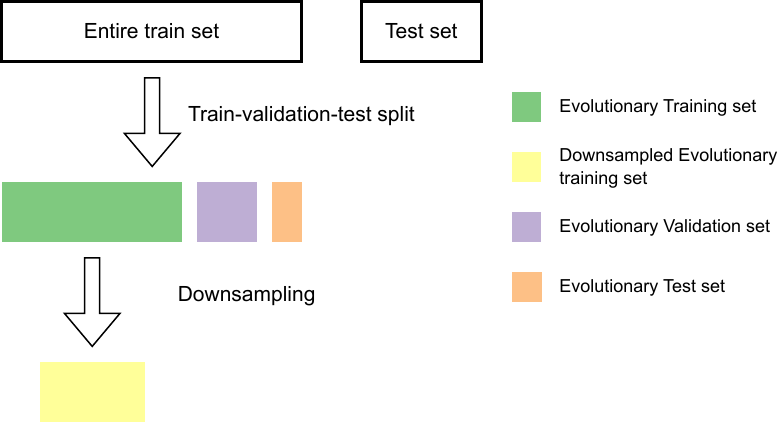}
  \caption{Set splits created by the dataset partitioning process. Coloured sets represent the ones used during the evolutionary process by supervised and/or self-supervised learning, whereas the test set is used at the end to check the final performance.}
  \label{fig:dataset_partitioning}
\end{figure}

There were two major factors taken into account while partitioning the dataset. The first one was that even though our fitness function is based on accuracy metric, we need to ensure that the data that is used to measure the fitness and the final performance of the networks are disjointed. For instance, if we were to use the same test set for fitness and performance purposes, even though individuals would not be trained on that data, the evolution process would still be guided by a measurement that is based on the test data. The other one was that the data partitioning process had to be flexible so that it could be used in supervised and self-supervised scenarios. Given that most common current benchmark datasets come divided in train and test sets, we used only data from the train set during the evolutionary process, whereas the test set was only used at the end of it, to check the performance of the best \gls{dnn} in unseen data. During the evolutionary process we divide the entire train set into three disjoint splits for training, validation and testing purposes. From now onwards, we will name these splits as evolutionary training set, evolutionary validation set and evolutionary test set. In order to test scenarios in which the amount of labelled data is scarce, we downsampled the train split further, producing an even smaller training subset which we will name downsampled evolutionary training set. All divisions and downsampling operations produce balanced sets, i.e., in each set there is the same number of instances per class. The entire dataset partitioning process is described in Figure~\ref{fig:dataset_partitioning}.

The splits that were used vary according to the learning paradigm chosen in the evolutionary process. In the supervised learning scenario, each candidate network is trained with the downsampled evolutionary training set. During the train of each individual, we use an evolutionary validation set to track network loss. This allows to set an early stopping criteria to preclude the network from overfitting. Finally, the evolutionary test set is used for fitness computation. 

\begin{figure}[ht]
  \centering
  \includegraphics[width=1.\linewidth]{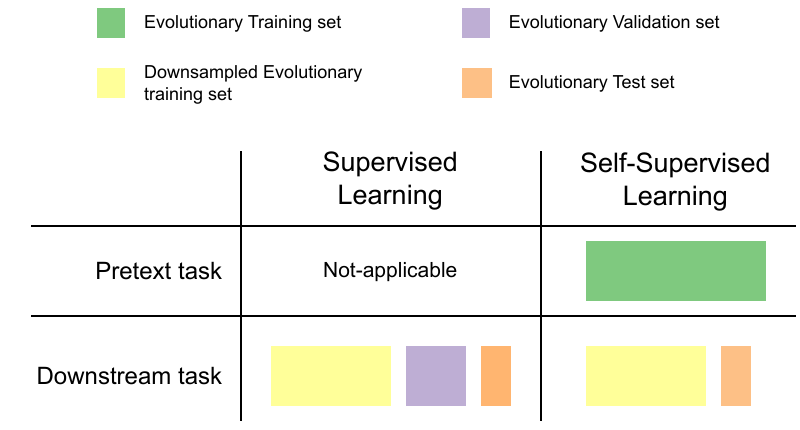}
  \caption{Set splits used breakdown by task and learning paradigm}
  \label{fig:partitioning_by_task}
\end{figure}

In the \gls{ssl} scenario, we use the evolutionary training set to train representations, without relying on its labels. Following previous \gls{ssl} works~\cite{zbontar2021barlow}, no evolutionary validation set is used in the pretext task, as it is harder to check if the network is overfitting. This happens because 1) the pretext task is not the final task we are aiming to solve, and 2) the loss produced during the pretext task is derived only from inputs and their stochastic augmented versions (no labels are used as ground truth). After representations are learned, the downstream task will make use of the downsampled evolutionary training set to train a network for the image classification problem to be solved. Finally, the evolutionary test set is applied to the trained network in order to compute the fitness of an individual. The splits that were used for each learning paradigm are described in Figure~\ref{fig:partitioning_by_task}.

\section{Experimentation}
\label{sec:experimentation}

The evolutionary engine built for these experiments (version 2.0.0 on GitHub) uses Python 3.9. As for the fitness assignment phase, Pytorch library was adopted to create the corresponding phenotype models and optimisers. Models were trained on the CIFAR-10 dataset~\cite{krizhevsky2009learning} using a single NVIDIA 3080 GPU.

\subsection{Setup}

In order to validate the adopted methodology, we conducted experiments that envisaged two goals. First, we wanted to analyse how would the evolution using a supervised learning paradigm compare against evolution using \gls{ssl}. Besides, we wanted to understand what was the impact of evolving \glspl{dnn} when the amount of labelled data is scarce. Therefore, four different scenarios were tested.
\begin{itemize}
    \item Evolution using supervised learning and 100\% of the labelled data for training;
    \item Evolution using supervised learning and 10\% of the labelled data for training;
    \item Evolution using Barlow Twins and 100\% of the labelled data for training (in the downstream task);
    \item Evolution using Barlow Twins and 10\% of the labelled data for training (in the downstream task).
\end{itemize}

For each of the aforementioned scenarios, 10 different runs were executed. Each run based on Barlow Twins took an average of 2 days to execute, whereas each of the supervised learning runs took less than a day. The parameters used for these experiments are described in Table~\ref{tab:setup}.

\begin{table}[ht]
\centering
\caption{Experimental parameters.}\label{tab:setup}
\begin{tabular}{c|cc}
\textbf{\gls{ea} Parameters} & \textbf{Supervised} & \textbf{\gls{ssl}} \\
\hline 
Number of runs & \multicolumn{2}{c}{10} \\
Generations & \multicolumn{2}{c}{75} \\
Population size & \multicolumn{2}{c}{6} \\
Add layer rate & \multicolumn{2}{c}{0.25} \\
Remove layer rate & \multicolumn{2}{c}{0.25} \\
DSGE mutation rate & \multicolumn{2}{c}{0.15} \\
Macro mutation rate & \multicolumn{2}{c}{0.3} \\
Train longer & \multicolumn{2}{c}{0.03} \\ \\
\textbf{Dataset Parameters} & \textbf{Supervised} & \textbf{\gls{ssl}} \\
\hline
Evolutionary Training Set & \multicolumn{2}{c}{70\%} \\
Evolutionary Validation Set & 20\% & - \\
Evolutionary Test Set & 10\% & 30\% \\ \\
\textbf{Learning Parameters} & \textbf{Supervised} & \textbf{\gls{ssl}} \\
\hline
$\lambda$ & - & 0.0078125 \\
Train time (mins) & \multicolumn{2}{c}{2} \\
Loss Function & Cross-Entropy & Barlow Twins\\
Downstream epochs & -  & 30 \\
Downstream learning rate & - & 0.001 \\
Downstream optimiser & - & Adam \\
Downstream loss & - & Cross-Entropy \\ \\ 
\textbf{Data Augmentation Parameters} & \textbf{Supervised} & \textbf{\gls{ssl}} \\
\hline
Padding & 4 & 4 \\
Horizontal Flip & 0.5 & 0.5 \\
Color Jitter (Brightness) & - & 0.4 \\
Color Jitter (Contrast) & - & 0.4 \\
Color Jitter (Saturation) & - & 0.4 \\
Color Jitter (Hue) & - & 0.1 \\
Color Jitter (Probability) & - & 0.8 \\
Grayscale & - & 0.2 \\
\end{tabular}
\end{table}

\subsection{Results}
 
\begin{figure}[ht]
  \centering
  \includegraphics[width=0.95\linewidth]{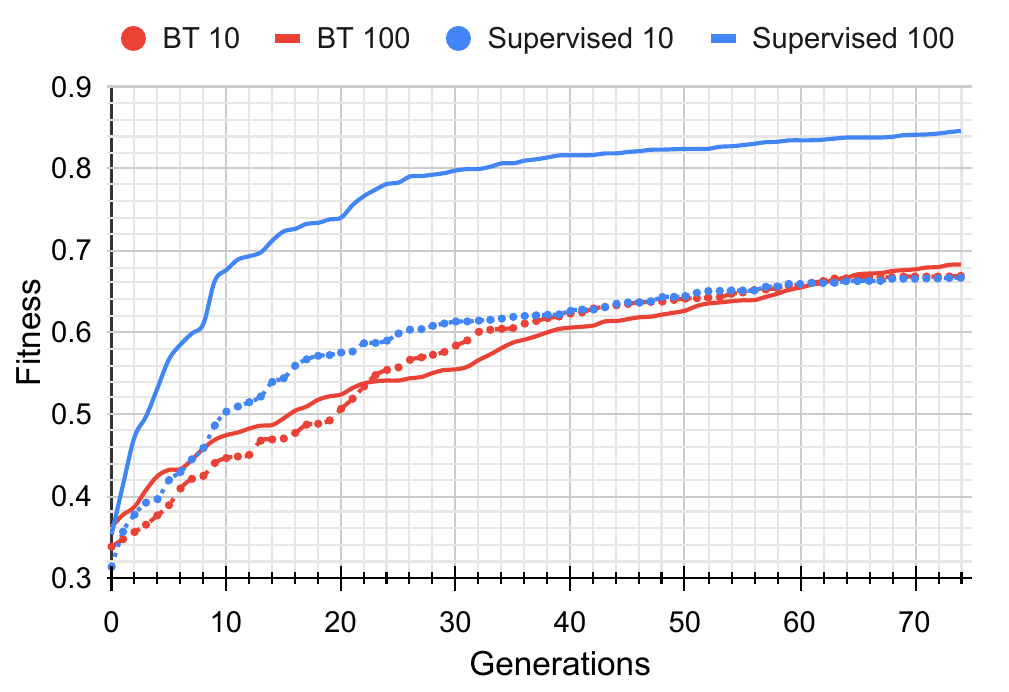}
  \caption{Evolution of \glspl{dnn} using the accuracy on the Evolutionary Test set as fitness. Results are averages of 10 runs.}
  \label{fig:evolution_plot}
\end{figure}

Firstly, we will focus in analysing the ability of the \gls{ne} algorithm to promote the emergence of increasingly better solutions as well as the generalisation capabilities of the best \glspl{dnn}. The plot in Figure~\ref{fig:evolution_plot} depicts how the fitness of the best individual changes throughout generations, averaged by the 10 runs.

From this Figure, it is clear that all four test cases are able to evolve solutions that are increasingly better. When the evolved networks learn with supervised learning and without downsampling (\textit{Supervised 100}), fitness converges faster towards more optimal regions as the blue line reaches an average fitness of 0.85. When compared to self-supervised runs that use Barlow Twins, this is a significantly higher fitness value. This behaviour is expected, as learning without labels makes the evolution task notoriously more difficult. Although the combination that produces solutions with the highest fitness is the \textit{Supervised 100} case, it has to be noted that when there is downsampling, the supervised scenario (\textit{Supervised 10}) seems to reach a plateau that suggests that no significant fitness improvements will happen if the evolution process was to be extended. Conversely, self-supervised based evolution exhibits evolution curves that suggest that further improvements are still possible to occur if the evolutionary runs are extended for more generations. This is particularly evident in the case where \glspl{dnn} are evolved without downsampling using a \gls{ssl} paradigm (\textit{BT 100}), as the evolution curve does not flatten at the same pace as the \textit{Supervised 100} line.

Another point worth mentioning is that \gls{ssl} guided evolution seems to be more resilient to the scarcity of labelled data than its supervised counterpart, given that both red lines are somewhat close to each other throughout generations, compared to the blue lines that represent supervised learning scenarios. This suggests that the evolved solutions are able to leverage unlabelled data to learn representations. Additionally, when the amount of labelled data is limited, self-supervised evolution is able to reach the same fitness levels as its supervised counterpart, despite the extra difficulty of the task.

Besides performing an evolution analysis, we computed the accuracy of the best individual of each run on the original test set. These values are summarised in Table~\ref{tab:accuracies} and they show the ability of the evolved networks to generalise on unseen data. When analysing the self-supervised runs, the best one is able to achieve a test accuracy of 77.41\%, using Barlow Twins with 100\% of labelled data. Moreover, the average performance of the best solutions evolved with \gls{ssl} was not affected by using fewer labelled samples in the downstream task, and it also obtained comparable results to the evolution runs guided by supervised learning using 10\% of labelled data, showing how competitive \gls{ssl} based evolution can be when the labelled data is limited.

\begin{table}[ht]
\caption{Test accuracies (\%) of the best networks produced by each testing scenario.}
\label{tab:accuracies}
\centering
\begin{tabular}{>{\raggedright\arraybackslash}p{0.5cm}|c|c|c|c}
 & \makecell{BT\\10\%} & \makecell{Supervised\\10\%} & \makecell{BT\\100\%} & \makecell{Supervised\\100\%} \\ \hline
Mean & 66.86 $\pm$ 3.53 & 66.53 $\pm$ 2.83 & 68.35 $\pm$ 5.73 & 84.67 $\pm$ 1.67 \\
\hline
\makecell{Best\\run} & 71.63 & 69.59 & 77.41 & 86.89 \\
\end{tabular}
\end{table}

\begin{figure}[ht]
  \centering
  \includegraphics[width=0.95\linewidth]{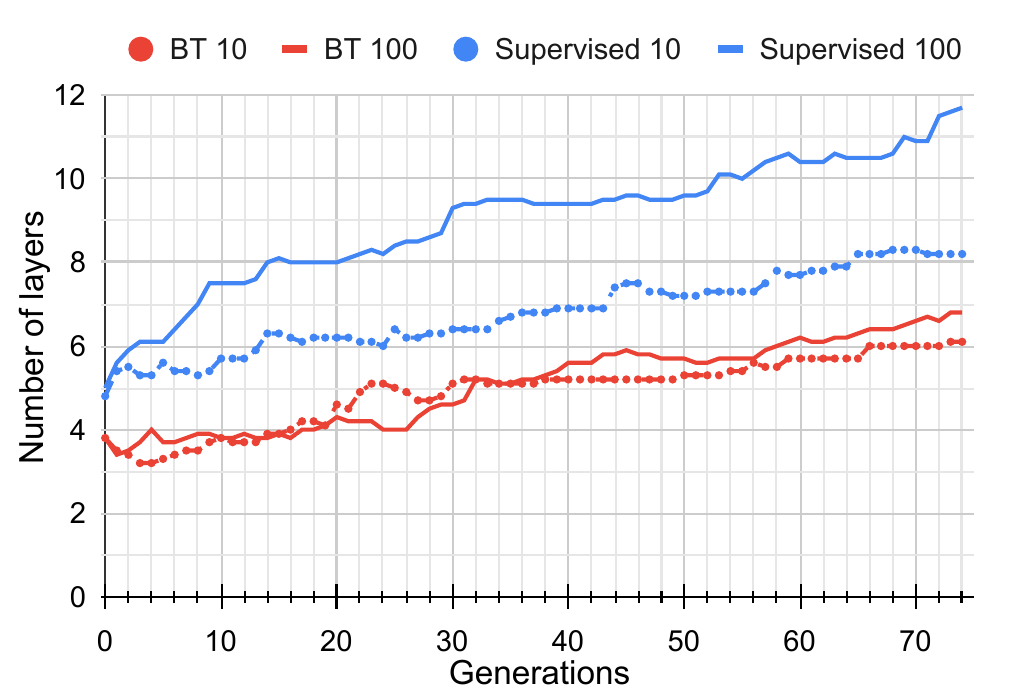}
  \caption{Number of layers of the best individuals throughout generations. Results are averages of 10 runs.}
  \label{fig:total_layers}
\end{figure}

The second part of our results involves a structural analysis to the best individuals whose average fitness was depicted previously. In order to do that, Figure~\ref{fig:total_layers} shows the average number of layers of the best individuals throughout generations. In the \gls{ssl} case, we consider the number of layers as the ones that are utilised to train representations and the final layer trained during downstream task, thus excluding the projector network.

Although we acknowledge that a direct comparison between the total number of layers of Barlow Twins and supervised scenarios cannot be made in a fairly manner, we can still analyse the impact of data scarcity in each of the learning paradigms separately. This is due to the fact that the downstream task is not evolved (only one fixed layer is used); hence, we do not promote the emergence of fully connected layers under the same circumstances. Nevertheless, we can still observe that evolved networks using Barlow Twins showed a similar number of total layers as the evolution proceeded, which can be seen by both red lines being very close to each other during the evolutionary process. That is not the case of the supervised scenario. The \textit{Supervised 100} line actually starts with a very similar average number of layers (5) than \textit{Supervised 10} (4.8). However, by the time of the last generation, there is a discrepancy in this number, as \textit{Supervised 100} reaches an average of 11.7 compared to 8.2 for the \textit{Supervised 10} scenario.

\begin{figure*}[!t]
  \centering
    \begin{subfigure}[ht]{0.42\linewidth}
        \centering
        \includegraphics[width=1.0\linewidth]{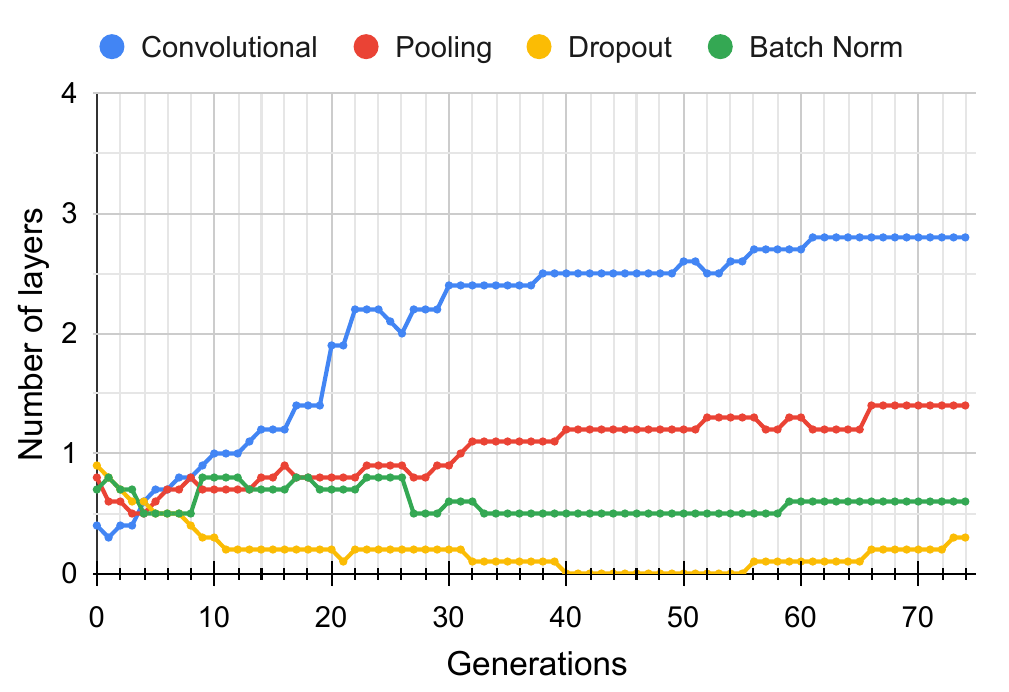}
    \caption{Barlow Twins with 10\% of labelled data.}
    \label{fig:bt_10_layers}
    \vspace*{0.5cm}
    \end{subfigure}
    \begin{subfigure}[ht]{0.42\linewidth}
        \centering
        \includegraphics[width=1.0\linewidth]{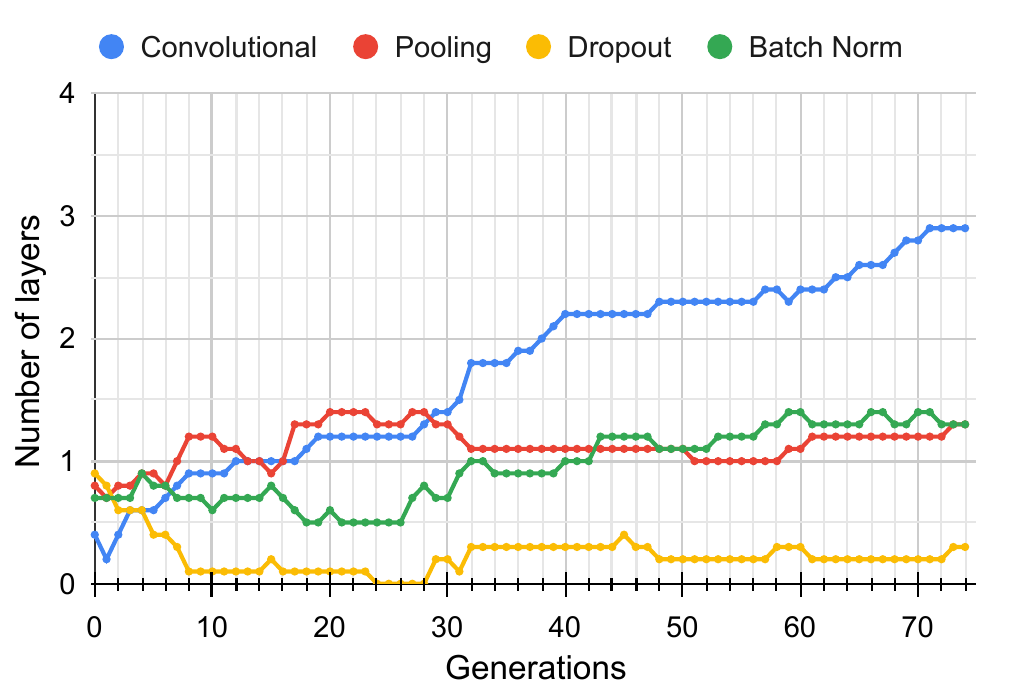}
        \caption{Barlow Twins with 100\% of labelled data.}
    \label{fig:bt_100_layers}
    \vspace*{0.5cm}
    \end{subfigure}
    \begin{subfigure}[ht]{0.42\linewidth}
        \centering
        \includegraphics[width=1.0\linewidth]{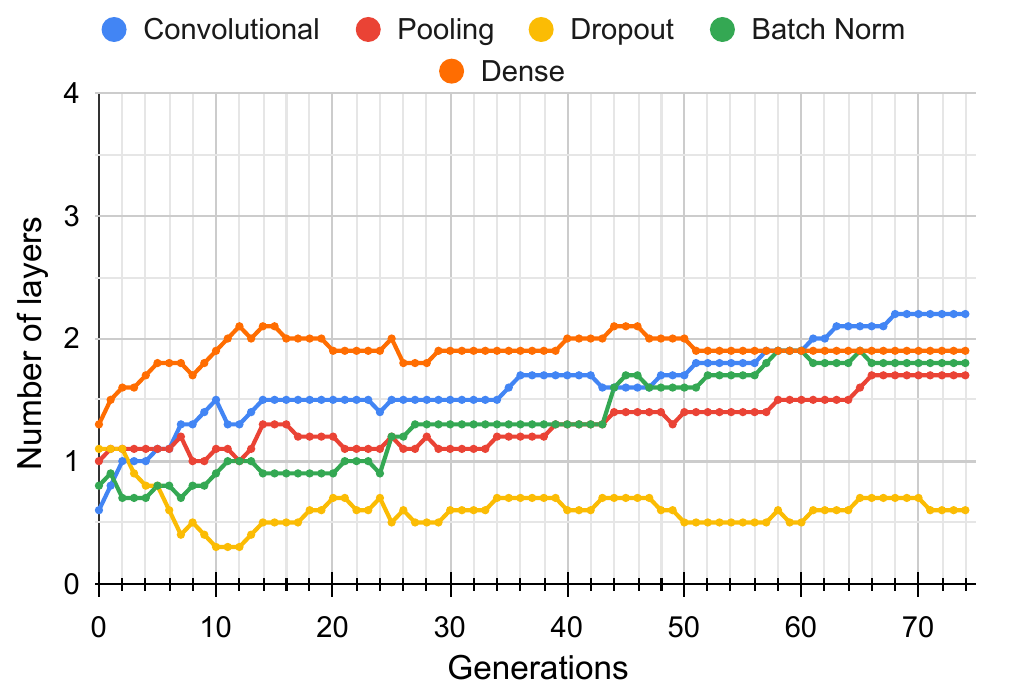}
    \caption{Supervised learning with 10\% of labelled data.}
    \label{fig:supervised_10_layers}
    \end{subfigure}
    \begin{subfigure}[ht]{0.42\linewidth}
        \centering
        \includegraphics[width=1.0\linewidth]{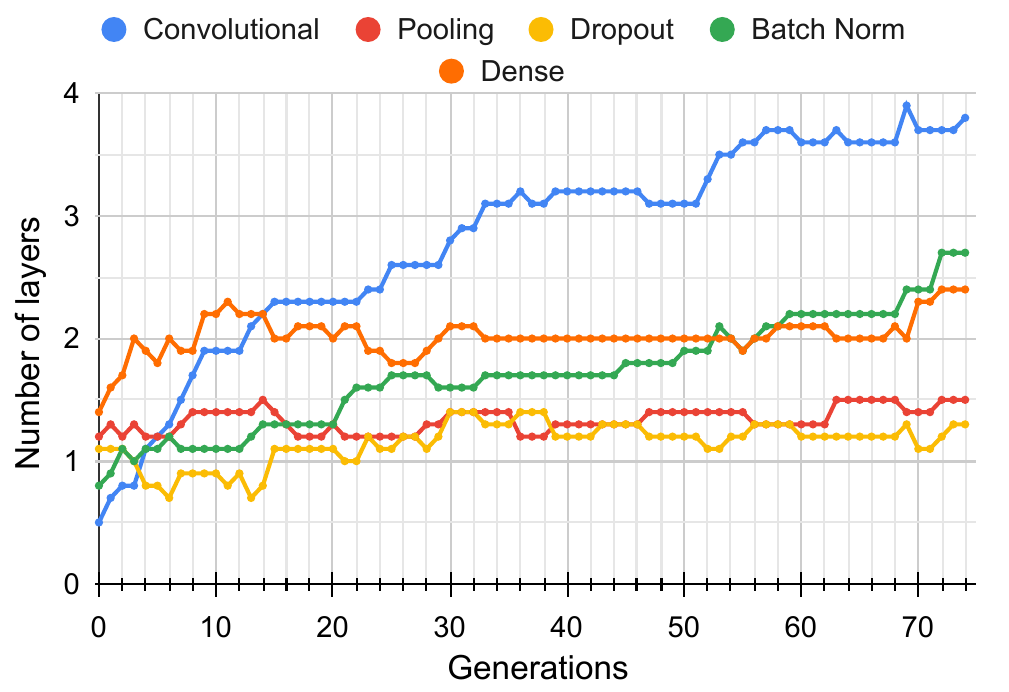}
        \caption{Supervised learning with 100\% of labelled data.}
    \label{fig:supervised_100_layers}
    \end{subfigure}
    \caption{Number of layers of the best individuals evolved using a different combination of learning paradigm and \% of labelled data. Results are averages of 10 runs.}
    \label{fig:layers_plots}
\end{figure*}

A deeper insight to the number of layers breakdown by type is depicted in Figure~\ref{fig:layers_plots}. This plot shows how there is a common trend when it comes to the number of convolutional layers. Regardless of the learning paradigm and percentage of labelled data used, the best individuals will tend to hold an increasingly number of convolutional layers throughout generations. This shows the importance of convolutional layers in promoting the emergence of increasingly better individuals.

Another detail that emerges from Figure~\ref{fig:layers_plots} is in the number of batch normalisation layers (green line). In Figures~\ref{fig:supervised_10_layers} and~\ref{fig:supervised_100_layers}, one can see that the green lines and blue lines (convolutional layers) grow together, suggesting that when the evolved networks learn through supervised learning, the number of batch normalisation layers has a trend that is correlated to the number of convolutional layers (even more visible in Figure~\ref{fig:supervised_10_layers}). On the other hand, if one performs the same analysis for Barlow Twins looking at Figures~\ref{fig:bt_10_layers} and~\ref{fig:bt_100_layers}, a different behaviour is observed (particularly in Figure~\ref{fig:bt_10_layers}), as the gap between the number of convolutional and batch normalisation layers seems to grow throughout generations. This is an interesting result, as the grammars used in supervised learning and Barlow Twins contain the same derivation rules for the definition of convolutional and batch normalisation layers.

\section{Conclusions and Future Work}
\label{sec:conclusions}

The time required to design \gls{dnn} architectures and to prepare the necessary datasets can be time-consuming. Moreover, in order to feed these models with data one needs to label them, which adds extra complexity to an already laborious task. This highlights the importance of reducing the amount of human intervention during this process.

The current work aims to alleviate this problem by combining ideas of \gls{ne} with \gls{ssl}, so that aspects of \gls{dnn} architecture design can be automated while reducing the reliance on labelled data to train these models. We developed a neuroevolutionary framework that evolves \glspl{dnn} which can learn either using supervised learning, or through Barlow Twins algorithm. In the latter, we leverage the unlabelled data to learn representations, and then train a single dense layer on top of them using a percentage of the labelled data. In order to assess the impact of the chosen learning paradigm within the \gls{ne} context, we evolve \glspl{dnn} that are taught with both supervised learning and \gls{ssl} paradigms. We also vary the amount of labelled data that is fed into the evolved networks (10\% or 100\%).

Our results confirm the idea that similarly to evolution guided by supervised learning, the same can be done in a self-supervised context. However, given the additional constraints imposed by the unavailability of labels, the optimisation of networks is a harder task. Despite of this, evolved networks that learn in a self-supervised manner are able to compete with their supervised counterparts when the amount of labelled data is limited, even outperforming them in some cases. Additionally, a structural analysis to the evolved \glspl{dnn} evidences that the ones that learned via the Barlow Twins algorithm tend to be structurally more similar (in terms of number of layers) regardless of the percentage of labelled data, when compared to networks that learned in a supervised manner.

In order to better understand the impact of \gls{ne} within a \gls{ssl} context, there are other components in our proposed framework which can be targeted by \gls{ec}. For instance, we plan to evolve other components like the projector network, as they are reported to help improving the performance of the downstream task. However, the reason why it works and its exact role are yet to be uncovered. Chen et al. concluded that using the layer before the projector as representations offered better downstream performance~\cite{chen2020simple}, whereas Ma et al. suggest that the projector seeks uniformity by maximising the distance between dissimilar samples~\cite{ma2023deciphering}. Similarly, we plan to evolve data augmentation aspects within Barlow Twins algorithm. They play a vital role during the pretext task by setting its difficulty to be solved. The network used during the downstream task can also be evolved to help the \gls{ea} to converge towards more optimal solutions.
On a different line of thought, we believe that modifying the search space by adopting a cell-based approach could benefit both the quality of the evolved networks and the time that takes to evolve them.

\section*{Acknowledgments}

This work is funded by the FCT - Foundation for Science and Technology, I.P./MCTES through national funds (PIDDAC), within the scope of CISUC R\&D Unit - UIDB/00326/2020 or project code UIDP/00326/2020

\bibliographystyle{ieeetr}
\bibliography{references.bib}

\end{document}